\documentclass[lettersize,journal]{IEEEtran}
\usepackage{graphicx}
\usepackage{xcolor, soul}
\sethlcolor{yellow}
\usepackage{amsmath,amsfonts}
\usepackage{algorithmic}
\usepackage{tabularx}
\usepackage{algorithm}
\usepackage{array}
\usepackage{textcomp}
\usepackage{stfloats}
\usepackage{url}
\usepackage{verbatim}
\usepackage{cite}
\usepackage{subfigure}
\usepackage{caption,setspace}
\usepackage{adjustbox}
\usepackage{threeparttable}
\usepackage{booktabs}

\usepackage[spaces,hyphens]{xurl} 
\usepackage[colorlinks,allcolors=blue]{hyperref}
\newcommand{\mycomment}[1]{}

\begin{document}

\title{Continuous Sign Language Recognition with Adapted Conformer  via Unsupervised Pretraining}
\author {Neena Aloysius, Geetha M, and Prema Nedungadi}


\markboth{Journal of \LaTeX\ Class Files,~Vol.~14, No.~8, August~2021}%
{Shell \MakeLowercase{\textit{et al.}}: A Sample Article Using IEEEtran.cls for IEEE Journals}

\maketitle

\begin{abstract}
Conventional Deep Learning frameworks for continuous sign language recognition (CSLR) are comprised of a single or multi-modal feature extractor, a sequence-learning module, and a decoder for outputting the glosses. The sequence learning module is a crucial part wherein transformers have demonstrated their efficacy in the sequence-to-sequence tasks. Analyzing the research progress in the field of Natural Language Processing and Speech Recognition, a rapid introduction of various transformer variants is observed. However, in the realm of sign language, experimentation in the sequence learning component is limited. In this work, the state-of-the-art Conformer model for Speech Recognition is adapted for CSLR and the proposed model is termed ConSignformer. This marks the first instance of employing Conformer for a vision-based task. ConSignformer has bimodal pipeline of CNN as feature extractor and Conformer for sequence learning. For improved context learning we also introduce Cross-Modal Relative Attention (CMRA). By incorporating CMRA into the model, it becomes more adept at learning and utilizing complex relationships within the data. To further enhance the Conformer model, unsupervised pretraining called Regressional Feature Extraction is conducted on a curated  sign language dataset. The pretrained Conformer is then fine-tuned for the downstream recognition task. The experimental results confirm the effectiveness of the adopted pretraining strategy and demonstrate how CMRA contributes to the recognition process. Remarkably, leveraging a Conformer-based backbone, our model achieves state-of-the-art performance on the benchmark datasets: PHOENIX-2014 and PHOENIX-2014T.
\end{abstract}


\section{Introduction}
\IEEEPARstart{S}{ign} language serves as the primary means of communication for the hearing-impaired, possessing its own distinct grammar, lexicon, and morphology. However, the majority of the hearing population lacks proficiency in this language. Consequently, effective communication between a deaf and hearing individual remains challenging and often necessitates the involvement of an certified interpreter. Nevertheless, the development of tools capable of automatically translating signs into speech or text, and vice versa, holds the potential to significantly diminish these communication barriers. The scope of such tools serves as a strong motivation for extensive research in the field of sign language, specifically within the realm of visual sign language. The primary visual tasks encompassed within this domain include continuous sign language recognition (CSLR) and sign language translation (SLT).

In recent years, significant advancements have been made in end-to-end CSLR systems utilizing neural networks. Recurrent neural networks (RNNs) have traditionally been the preferred choice for CSLR, as they effectively model the temporal dependencies within gesture sequences [\cite{hu2023self}, \cite{hao2021self}, \cite{koishybay2021continuous}]. However, the transformer architecture, based on self-attention mechanisms, has gained widespread acceptance for sequence modeling due to its capacity to capture long-range interactions and efficient training [\cite{camgoz2020sign}, \cite{niu2020stochastic}, \cite{zhou2021signbert},  \cite{zhou2022cross}, \cite{aloysius2021incorporating}]. Additionally, convolutional approaches have also demonstrated success in CSLR by progressively capturing local context through layered receptive fields [\cite{hu2023self}, \cite{guo2019dense}, \cite{cheng2020fully}]. Nevertheless, both self-attention models and convolutional models have their respective constraints. Transformers excel in modeling extensive global context but struggle to capture intricate local feature patterns. Conversely, Convolutional Neural Networks (CNNs)  excel at leveraging local information and serve as a fundamental computational block in visual tasks. They acquire knowledge through shared position-based kernels over a localized window, maintaining translation equivariance, and effectively capturing features such as edges and shapes. Nevertheless, a drawback of depending solely on local connectivity is the need for a greater number of parameters or layers to capture global information. This challenge calls for a model that successfully integrates both convolution and self-attention mechanisms, exemplified by the state-of-the-art Automatic Speech Recognition (ASR) model, known as the Conformer \cite{gulati2020conformer}.

This work adapts the Conformer architecture as ConSignformer for the purpose of CSLR. A multi-modal network of RGB and heatmap videos was pursued, employing CNNs as the feature extractor. Attentional Pyramid Network (APN), a new pyramid network was incorporated into the training network, to enhance the training of the multi-modal pipeline architecture. APN was removed during the inference stage.  During continuous sign gesturing, the self-attention mechanism in the Conformer was replaced with a novel Cross-Modal Relative Attention (CMRA), to enhance context learning, which involved swapping of query vectors between the pipelines. In order to reinforce the Conformer and provide a robust weight initialization, an unsupervised pretraining step was introduced. Close scrutiny of the recent successful transformer models, expound  that pretraining significantly improves fine-tuning on downstream tasks. Apparently, this work marks the first instance of pretraining the Conformer architecture. The unsupervised pretraining approach employed equips the model with valuable priors, rendering it capable of performing effectively in diverse sign language related tasks.  After the pre-training phase, we meticulously craft straightforward yet highly efficient prediction heads for continuous sign recognition. These prediction heads are then fine-tuned in conjunction with the pre-trained Conformer encoder to tailor them to specific downstream tasks.

In summary, our contributions are as follows:
\begin{itemize}
\item ConSignformer, a multi-modal pipelined framework, spawned from Conformer model adapted for   CSLR. 
\item Regressional Feature Extraction- unsupervised pretraining of the Conformer, performed on curated continuous Indian Sign Language (ISL) pose data with Mean Squared Error loss.
\item Sign Language agnostic Conformer model, pretrained on ISL, practicable for diverse tasks, in the sign language domain (e.g., ISL, ASL, AUSLAN).
\item Innovative Attentional Pyramid Network for enhanced training of ConSignformer 
\item The relative attention of Conformer converted to Cross-Modal Relative Attention (CMRA) enables contextual learning of gestural sequences.

\item Extensive experiments conducted to validate the feasibility and effectiveness of the proposed framework. The results demonstrate state-of-the-art WER on the German datasets Phoenix 2014 and Phoenix 2014T.
\end{itemize}

\section{Related Works}
\label{sec:related}
In this section, we commence by providing a literature review for continuous sign language recognition based on video inputs. Following that, we study the significant transformer variants spanning various domains.

\subsection{Continuous Sign Language Recognition}
The study begins with our literature review \cite{aloysius2020understanding} in the field of sign recognition and highlights recent advancements in CSLR.  The Deep Learning based CSLR process involves three core modules: feature learning module for encoding short-term spatial details, sequence learning module that captures long-term context and alignment learning module for ensuring accurate synchronization between video clips and glosses for reliable training. Common models for feature extraction include 2D CNNs [\cite{hao2021self},\cite{min2021visual}, \cite{zhou2020spatial}, \cite{hu2023self}], 3D CNNs [\cite{zhou2019dynamic}, \cite{pu2019iterative}] and body keypoints extractors [\cite{jiang2021skeleton}, \cite{chen2022two}, \cite{aditya2022novel}]. For sequence learning, RNNs [\cite{koishybay2021continuous}, \cite{hu2023self}, \cite{hao2021self} ], 1D CNNs [\cite{hu2023self}, \cite{guo2019dense}, \cite{cheng2020fully}], and more recently, Transformers are employed [\cite{camgoz2020sign}, \cite{niu2020stochastic}, \cite{zhou2021signbert},  \cite{zhou2022cross}, \cite{aloysius2021incorporating}, \cite{hu2023signbert+}]. Although HMM-based alignment and decoding techniques [\cite{koller2019weakly}, \cite{koller2018deep}] were initially utilized, encoder-decoder based attention networks [\cite{huang2018video}, \cite{guo2018hierarchical}, \cite{camgoz2018neural}] were later experimented but now Connectionist Temporal Classification (CTC) \cite{graves2006connectionist} decoding [\cite{zhou2020spatial}, \cite{zheng2023cvt}, \cite{hao2021self}, \cite{camgoz2020sign}, \cite{chen2022two}] has become the prevalent method.

The best results are achieved with multi-modal networks [\cite{chen2022two}, \cite{chen2022simple}, \cite{zhou2020spatial}, \cite{jiang2021skeleton}], even though fusion techniques introduce complexity and noise susceptibility. Multi-modal mainly refers to the raw RGB videos and the body pose estimates or human keypoints. The keypoints are directly regressed from the video frames as in Deep Pose \cite{toshev2014deeppose} and Mediapipe \cite{mediapipe} or estimated as heatmaps as in the case of HRNet \cite{wang2020deep}, OpenPose \cite{cao2017realtime} and other similar networks \cite{tompson2014joint}, \cite{newell2016stacked}. The Two-Stream model proposed by Chen et al. \cite{chen2022two} is the current state-of-the-art in CSLR. It comprises knowledge distillation, multiple auxiliary losses, and Spatial Pyramid Networks that are adeptly utilized to compensate for the data scarcity. The pretraining strategy proposed in their previous work \cite{chen2022simple} has also been beneficial in their subsequent work.

Single cue models [\cite{camgoz2018neural}, \cite{camgoz2020sign}, \cite{yin2020attention}, \cite{hu2023self}, \cite{aloysius2021incorporating}, \cite{koishybay2021continuous}] maintains a trade off between model complexity and performance. An efficient approach is utilizing a single cue with cross-modal alignment, as seen in CVT-SLR \cite{zheng2023cvt}, which employs a Variation-AutoEncoder for sequence learning. The SMKD approach \cite{hao2021self} combines a single cue and 2D CNN-BiLSTM-CTC in its recognition network, using a three-stage optimization strategy to reduce error rates. The most recent SignBERT+ model \cite{hu2023signbert+} has introduced an innovative pretraining technique for the BERT architecture, integrating hand-prior information for the recognition task, which has demonstrated significant benefits.

Iterative training is another approach adopted for optimized learning and to mitigate the concern of overfitting. It was initially proposed by Koller et al. \cite{koller2016deep}. This approach was further improved in \cite{koller2017re} and the recent works \cite{cui2019deep}, \cite{pu2019iterative}, \cite{camgoz2018neural} and \cite{camgoz2020sign} have also adopted the technique. Such training serves to boost the feature extractor. The Visual Alignment Constraint (VAC) \cite{min2021visual}  proposed by Yuecong et al. goes even further by enhancing the feature extractor through alignment supervision. This approach introduces a novel viewpoint on how the visual and alignment modules interrelate. Differently, Hu et al. \cite{hu2023self} enhanced the feature extractor by adding weightage to informative spatial features by proposing self-emphasizing network. While these methods enhance the effectiveness of the spatial module, the sequence learning module does not derive as much benefit. Masked Iterative Training (MAIT) \cite{zhou2022cross} is one such technique that empowers not only the visual module but also the sequence learning module, especially when BERT is employed. Additionally, it improves the overall robustness of the system, and the utilization of random masking helps augment the variety of training data.

\subsection{Transformer Variants}
Transformers \cite{vaswani2017attention} have excelled in AI across multiple domains, including natural language processing (NLP),  audio processing and computer vision. Even in the sign language domain, many works related to SLT tasks began to emerge with the introduction of the Transformer model. Given this context, it's quite natural that more researchers have been drawn to this area. As a result, there are now various Transformer variants available across different domains. Researchers also explore the adaptation of variants initially developed in one domain to excel at related tasks in other domains. For instance, the well-known BERT architecture \cite{kenton2019bert} was originally designed for NLP, but we can observe its adaptation for vision [\cite{zhou2021signbert}, \cite{tunga2021pose}, \cite{kalfaoglu2020late}] and speech tasks [\cite{huang2021speech}, \cite{chung2021w2v}, \cite{chiu2021innovative}, \cite{d2020bert}] as well.

The Transformer is an encoder-decoder architecture. Depending on the application, either the encoder, the decoder, or both can be utilized. Tasks like classification, sequence labeling, and the recognition of action or gesture sequences typically make use of the Transformer's encoder component. Widely used encoders include BERT \cite{kenton2019bert}, RoBERTa \cite{liu2019roberta}, and BigBird \cite{zaheer2020big}. For language modeling, the Transformer decoder and its variants see extensive use, with some notable examples being GPT \cite{radford2018improving}, GPT-2 \cite{radford2019language} and GPT-3 \cite{brown2020language}. In machine translation tasks such as SLT, speech translations, and text-to-text translations, both the encoder and decoder are employed. Frequently used encoder-decoder Transformer models include BART \cite{lewis2019bart}, T5 \cite{raffel2020exploring}, and Switch Transformer\cite{fedus2022switch}, among others. A detailed survey of the major architectures are outlined in the study by  Lin et al. \cite{LIN2022111}. It reviewed the various X-formers based on architectural variations, pretraining and its applications. 

ASR and CSLR are essentially dealing with closely related problems, except for the input types. The current state-of-the-art speech recognition model is Conformer-1 \cite{conformer1}, released by Assembly AI. Its initial variant was Conformer \cite{gulati2020conformer}, developed by Google Brain, which marked the first time self-attention and convolution were combined for effective global and local learning while being a size-efficient architecture. This model significantly reduced error rates in speech recognition. The main drawback was its complexity in terms of memory and computation, which made the original Conformer architecture slower to operate in both training and inference tasks compared to other existing architectures. Subsequent research aimed to reduce the complexity without compromising performance, resulting in an upgraded version called Efficient Conformer \cite{burchi2021efficient}. This development paved the way for a production-ready ASR system, and the Efficient Conformer was further modified to enhance performance on noisy data, culminating in the creation of Conformer-1.

Building on our literature review above, we were inspired to adapt the Conformer model for sign recognition task. We begin our investigation with the original Conformer model and aim to incrementally enhance it, following a similar approach to the evolution of Conformer-1. Given its proficiency in modeling both local and global dependencies, which is crucial for sign recognition, experimentations with this model could yield valuable results and potentially pave the way for a robust sign recognition system. The main challenge in adapting the model lies in designing an efficient attention module that meets our specific requirements. 
\begin{figure}
    \centering
    \subfigure[Mediapipe Pose. Only upper body estimates were considered that is included in gray region.]{\includegraphics[width=0.35\textwidth]{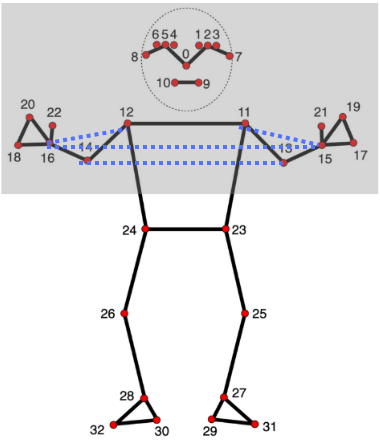}}

    \subfigure[Mediapipe Hands]{\includegraphics[width=0.49\textwidth]{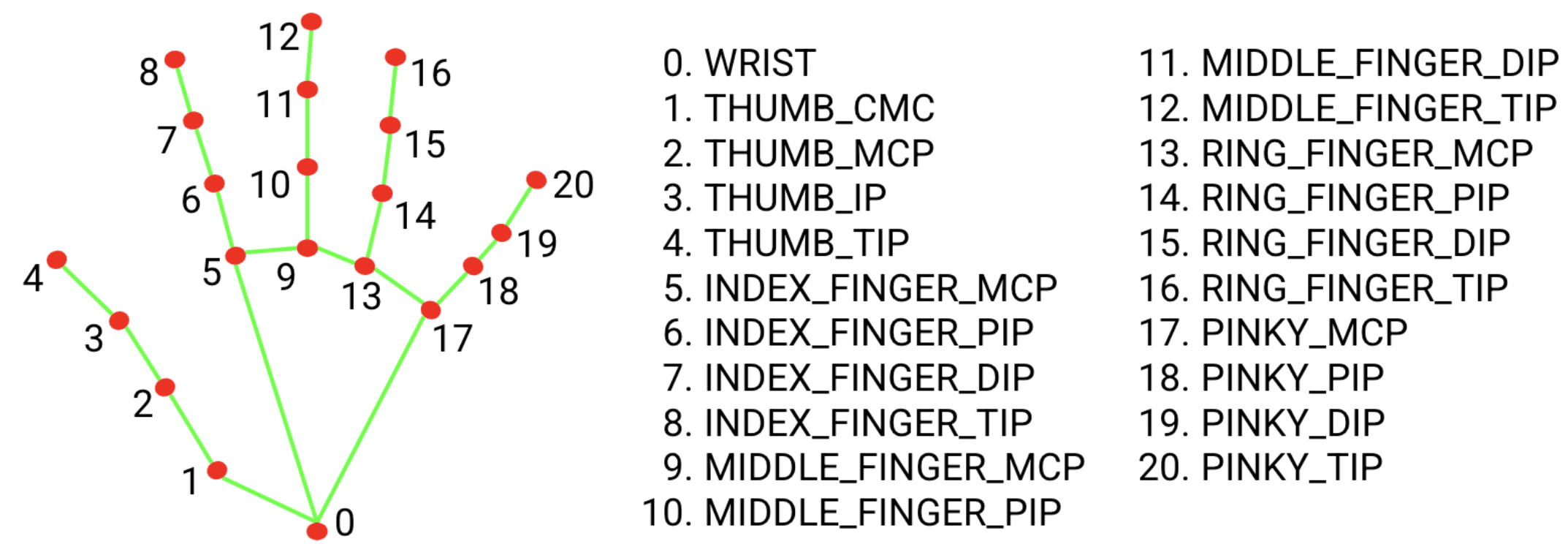}}

    \caption{Joint estimates of (a)mediapipe pose and the euclidean distance between joints shown in blue lines (b) mediapipe hands.  \tiny{[Adapated from \cite{mediapipe}}]}
    \label{keypoints}
    
\end{figure}

\section{Proposed Method}

Our proposed model is an ensemble of a hybrid network composed of a CNN model called S3D and a Transformer variant called Conformer. Since this is a complex deep network, accurate training demands a large amount of annotated data, which, in this case, consists of sign language gesturing videos. In such scenarios, the use of pretrained models and transfer learning has been beneficial for the downstream task, allowing training with relatively fewer data than would otherwise be required.

Our work began with pretraining the Conformer model on curated isolated gesture word videos and using the transfer-learned weights of S3D, which was trained on the Kinetics-400 action dataset. The hybrid network was then pretrained individually with CTC loss using continuous RGB and heatmap videos. Finally, ensemble training was conducted with the two pretrained networks and a third Conformer pipeline, that works on fused features from the two S3D models, under the supervision of CTC.
\subsection{Unsupervised Pretraining}
Pretraining is a common practice for resource-intensive models such as CNNs and Transformer variants like BERT and mBART. It serves as an effective method for initializing model weights. When pretraining is done in an unsupervised manner, there is the added advantage of avoiding the cost of labeling a vast dataset. Notably, pretraining yields better results when there is a degree of similarity between the pretraining data and the downstream task data. This approach enables the model to acquire specific data features in advance, essentially establishing prior knowledge that benefits the subsequent task. The choice of the pretraining task also plays a vital role in determining the effectiveness of learning these priors. This work proposes a novel pretraining methodology that prioritizes the trajectory of gestures and enhances the model's comprehension of the salient regions in both pretrained videos and videos in the fine-tuned task.

In this context, our objective was to develop a well-pretrained Conformer model for SLR. This effort was part of a significant project supported by the Government of India, focused on Indian Sign Language (ISL). We amassed an extensive dataset comprising nearly 12,000 videos, recorded by deaf signers and certified ISL interpreters. Utilizing Mediapipe Pose, we extracted 22 upper body keypoints for each video frame. The highlighted portion of Figure \ref{keypoints}(a) shows the keypoints of our interest. 
These keypoints represent 3D joint estimations, including spatial coordinates (X, Y) and depth measurement (Z). Additionally, we derived several supplementary features from these joint estimates:
\begin{enumerate}
    \item Bone Length: Calculated as the Euclidean distance between connected pose joints.
    \item Joint Relative Position: Describes the relative positioning of each joint in relation to its adjacent or connected joints.
    \item Joint Euclidean Distances: Includes distances such as left hand to right hand palm, left hand to right hand elbow, left palm to left shoulder, and right palm to right shoulder distances. The distances are marked in blue lines in Figure \ref{keypoints}(a).
\end{enumerate}

 Likewise, by utilizing Mediapipe Hands as in Figure \ref{keypoints}(b), 
 we extracted 21 3D joint estimations for each hand. Additional features, including bone length, joint relative positions, and Euclidean distances between specific joints (8,0), (12,0), (16,0), (20,0), and (4,0), were also computed.

\begin{figure*}
 \centering
    \includegraphics[scale = 0.9]{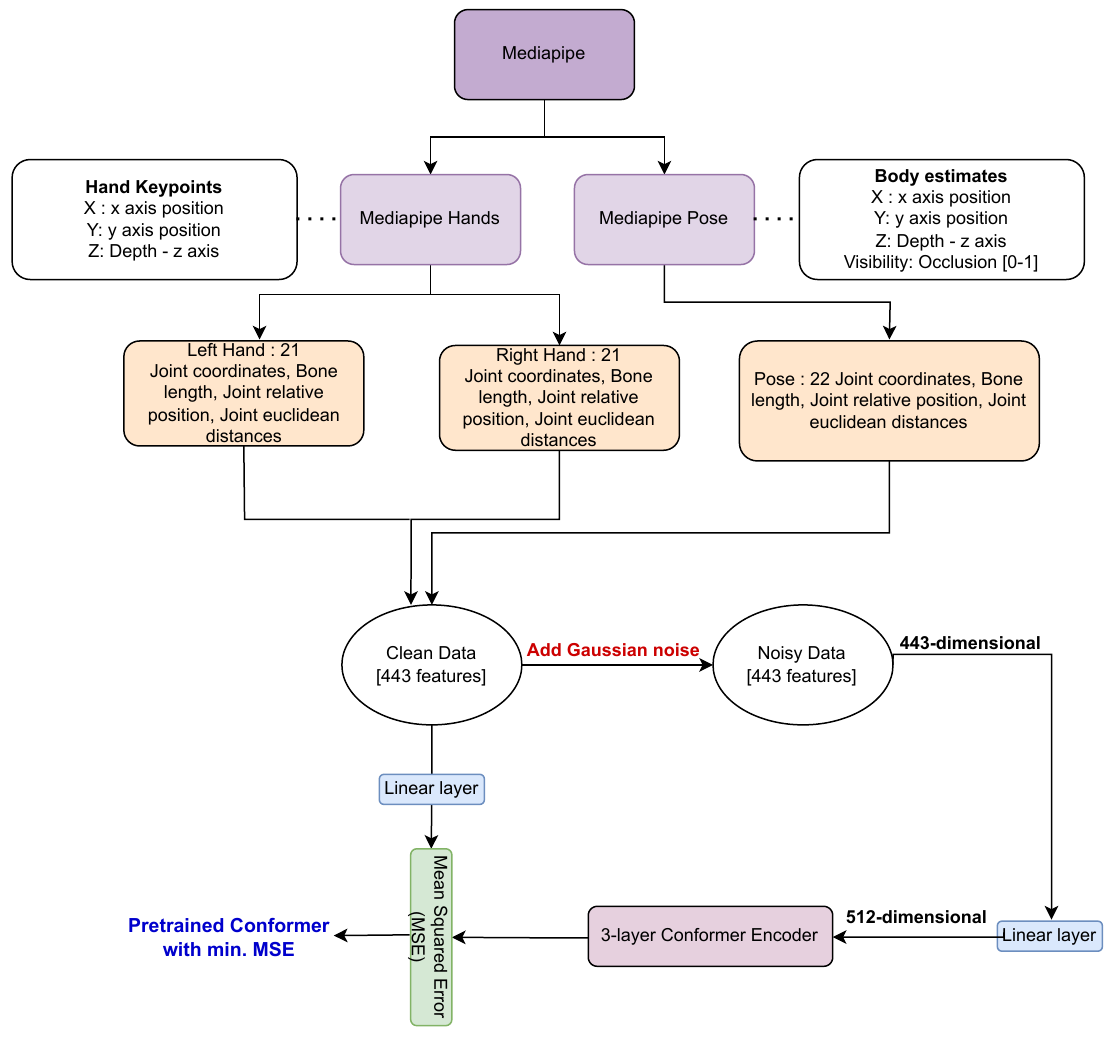}
    \caption{Unsupervised Pretraining - Regressional Feature Extraction}
    \label{regression}
\end{figure*}
The keypoints are combined to form a 443-dimensional vector. This dataset is split into two parts: (i) clean data, consisting of the original keypoints, and (ii) noisy data, where Gaussian noise with a standard deviation of 0.2 is applied. Both the clean and noisy data are then passed through a linear layer to be transformed into a 512-dimensional vector, aligning with the hidden dimension of the Conformer model. The noisy data is fed into the model, and our expectation is for the model to predict the clean data. This process is supervised by applying a Mean Squared Error loss between the two datasets. If we denote:
\begin{itemize}
    \item $X$ as clean gesture keypoints
    \item $X_{noisy}$  as the noisy version of the gesture keypoints, which is the input to your model
    \item $X_{predicted}$ as the output of your model, which should ideally be a denoised version of $X_{noisy}$
\end{itemize}
 then, the MSE loss can be defined as: \newline
 \begin{equation}
MSE\ Loss = \frac{1}{N} \sum_{i=1}^{N}  \| X_i - X_{predicted,i} \| ^2     
 \end{equation}
 where $N$ is the number of samples.

We refer to this novel unsupervised pretraining technique as 'Regressional Feature Extraction' which is depicted in Figure \ref{regression}. The Conformer model, once trained on the ISL dataset, serves as a pretrained sequence learning model applicable to various sign language-related tasks across different domains.
\begin{figure*}
    \centering
    \includegraphics[scale = 0.72]{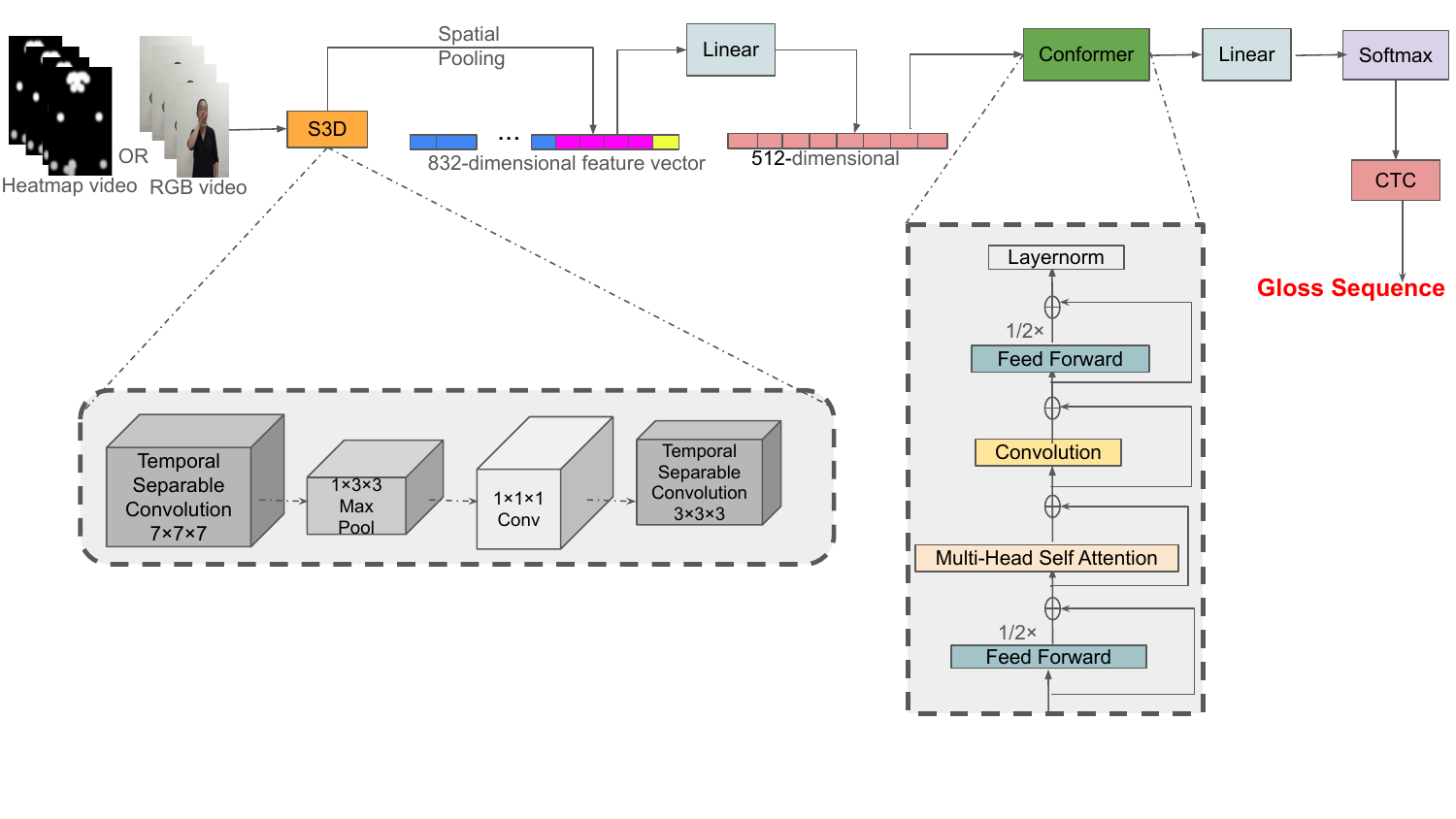}
    \caption{RGB Encoder - when input is RGB videos; Heatmap Encoder - when input is heatmap videos. Heatmaps are the representations of the keypoints of the face, hands, and upper body, extracted using HRNet \cite{jin2020whole} trained on COCO-WholeBody \cite{wang2020deep}.}
    \label{rgb/heatmap}
\end{figure*}

\subsection{ConSignformer}
Our proposed model is an ensemble of three networks: the RGB Encoder, Heatmap Encoder, and the Fusion Network, collectively referred to as 'ConSignformer.' The fundamental architecture is consistent across these three ensembles, consisting of a CNN as the feature extractor and the  pretrained Conformer as the sequence learning module. The distinction among them lies in the type of input they process. Specifically, the RGB Encoder operates on RGB videos, the Heatmap Encoder operates on heatmaps, and the Fusion model operates on features fused from both RGB and Heatmap inputs. Both the Heatmap and RGB encoders are pretrained separately in a task-adaptive manner. Subsequently, they are jointly trained with the Fusion network in an end-to-end fashion. The following sections provide detailed explanations of the RGB and Heatmap Encoders, including their pretraining strategies. This is followed by a discussion of the Fusion network, and finally, an in-depth exploration of the proposed ensemble model, ConSignformer.

\subsubsection{RGB/Heatmap Encoder}
Figure \ref{rgb/heatmap} illustrates the fundamental CNN-Conformer architecture, trained under the supervision of the Connectionist Temporal Classification (CTC) loss. Feature extraction is performed using the state-of-the-art CNN model, S3D, which has been pre-trained on the Kinetics-400 dataset \cite{kay2017kinetics}. The parameter-efficient design of the layers and the decoding speed prompted us to choose this model. The representation from the first four layers sufficed for our purpose of obtaining dense spatio-temporal video feature representations. These representations are spatially pooled to obtain 832-dimensional video embeddings, which are further passed through a linear layer to reduce them to 512 dimensions to match the Conformer hidden dimension. In contrast to normal transformers, here the embeddings are directly passed to the Conformer without the addition of any positional information. This is because the Conformer has a relative position embedding mechanism inside its attention heads. When the S3D-Conformer works on RGB videos, it is referred to as the RGB-Encoder, and when it works on heatmaps, it is called the Heatmap-Encoder. Each encoder is separately pretrained on the German datasets, under CTC supervision. This is a task-adaptive pretraining designed to adapt the model to the task of CSLR. 
\begin{figure*}
    \centering
    \includegraphics[scale = 1]{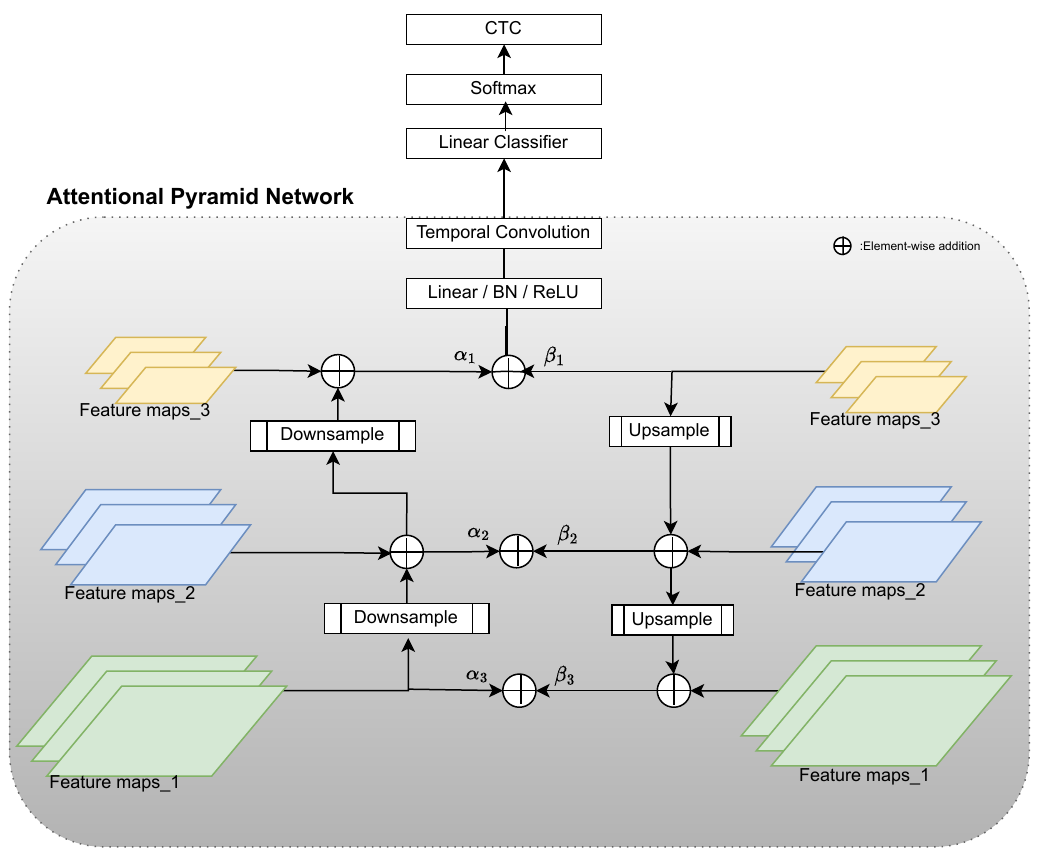}
    \caption{Attentional Pyramid Network. $Feature\ maps\_i$ is the intermediate representations of S3D. The left side shows the Top-Down flow and on the right side is the Bottom-Up Flow with lateral connections to result in a Parallel Flow of features. Attention weights are denoted by $\alpha_i$ and $\beta_i$}.
    \label{apn}
\end{figure*}

\subsubsection{Fusion Network}
The 512-dimensional spatial embeddings from the S3Ds of the Heatmap and RGB encoders were combined through element-wise addition to yield fused features. The S3D-Conformer that operates on these fused features is referred to as the Fusion Network. This setup does not undergo any pretraining procedure. Instead, it serves as an ensemble along with the two pipelines mentioned above to enhance the overall performance of sign recognition during joint training.

\begin{figure*}
    \centering
    \includegraphics[scale = 0.71]{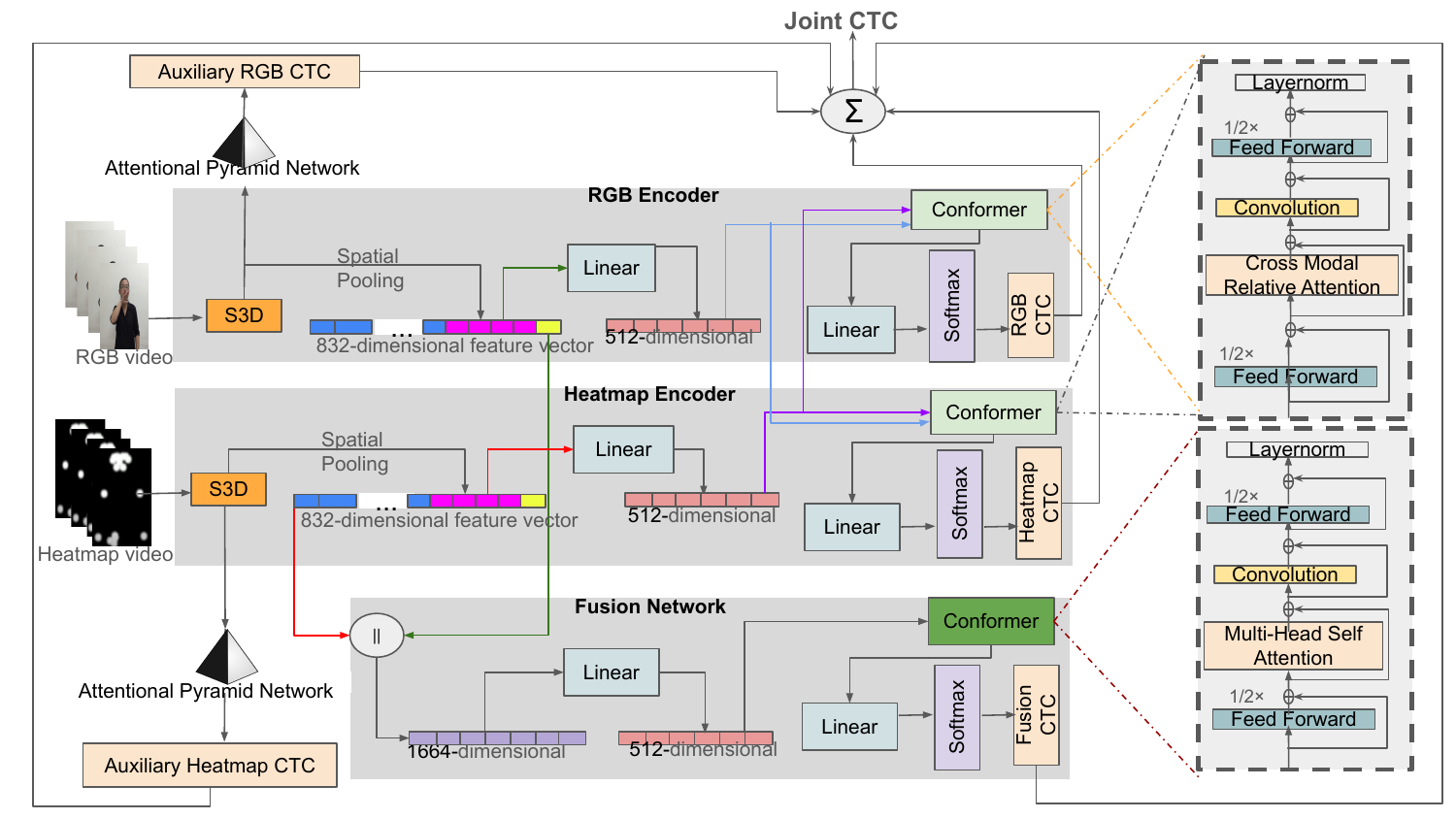}
    \caption{ConSignformer - Ensemble of RGB Encoder, Heatmap Encoder and Fusion Network, supplemented by Attentional Pyramid Networks.}
    \label{proposed}
\end{figure*}
\subsubsection{Attentional Pyramid Network}
The Temporal Pyramid Network, as proposed by Yang et al. \cite{yang2020temporal}, is designed to integrate at the feature level with different 2D or 3D backbone networks. We have adapted this network by following the parallel flow of feature fusion and introduced an attention mechanism to learn the weights assigned to important features. Therefore, we have named this network the Attentional Pyramid Network (APN), as depicted in Figure \ref{apn}. The combined features are then temporally convolved after transformation into a 512-dimensional, normalized vector. The Conformer is not integrated into APN to maintain simplicity. As an auxiliary task in the final joint ensemble training for CSLR, APN is incorporated into the RGB and Heatmap pipelines. This network is also trained using the CTC loss, which enhances its effectiveness in sign language recognition. 

\subsubsection{Ensemble}
The final ensemble model, composed of the RGB Encoder, Heatmap Encoder, and Fusion Network, is referred to as the ConSignformer, as shown in Figure \ref{proposed}. APN is added as auxiliary networks to the RGB and Heatmap pipelines. The auxiliary networks supplement the training and is not a part of the ConSignformer in the inference stage.  

In the ensemble setup, the relative multi-head attention of the Conformers in the RGB and Heatmap encoders works differently. The output of the Multi-Head Attention (MHA) module is generated as usual by concatenating the representations learned by all the heads.

\begin{equation}
    attention(Q,K,V)\ \textbf{=}\ softmax \left(\frac{QK^T} {\sqrt{d_z}}\right)V, 
\end{equation}
where \textbf{Q}, \textbf{K}, \textbf{V}  are the query, key and value matrices generated from the input set of vectors using three different transformations $W^q,W^k,W^v$ respectively.  This is the normal self-attention with relative position embedding as used in the Fusion network. The attention in the Heatmap/RGB pipeline is modified to accept the query as the RGB/Heatmap input embedding, and vice versa. Accordingly, equation (2) becomes:

\begin{equation} 
    \begin{aligned}
        attention_{rgb}\ &\textbf{=}\ attention(Q_{rgb},K_{rgb},V_{rgb})\  \textbf{+}\ \\
                        &\quad  attention(Q_{hm},K_{rgb},V_{rgb})
    \end{aligned}
\end{equation}
and
\begin{equation}
 \begin{aligned}
    attention_{hm}\ &\textbf{=}\ attention(Q_{hm},K_{hm},V_{hm})\  \textbf{+}\ \\
     &\quad  attention(Q_{rgb},K_{hm},V_{hm})
    \end{aligned}
\end{equation}
where $hm$ denotes heatmap.
This new modified attention mechanism is known as Cross-Modal Relative Attention (CMRA). It facilitates interaction between the two pipelines, thereby aiding invaluable information exchange. The attention mechanism in the Fusion model remains unchanged.

The cumulative loss of ConSignformer, denoted as $L_{CSLR}$, encompasses the aggregation of individual losses of the RGB Encoder ($L_{rgb}$), Heatmap Encoder ($L_{heatmap}$), Fusion Network ($L_{fusion}$) and the two auxiliary losses originating from APNs ($L^{\small{rgb}}_{\small{apn}}$ and $L^{\small{heatmap}}_{\small{apn}}$). ConSignformer jointly optimizes the total loss as follows:

\begin{equation}
   L_{CSLR}\ =\ L_{rgb}\ \textbf{+}\ L_{heatmap}\ \textbf{+}\ L_{fusion}\ \textbf{+}\ L^{\small{rgb}}_{\small{apn}}\ \textbf{+}\ L^{\small{heatmap}}_{\small{apn}}
\end{equation}

\section{Experiments}
\subsection{Dataset}
\begin{figure}
    \centering
    \subfigure[Train data]{\includegraphics[width=0.3\textwidth]{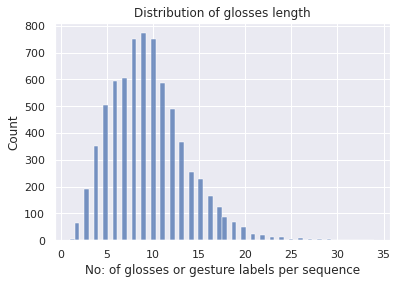} }
    \subfigure[Validation data]{\includegraphics[width=0.3\textwidth]{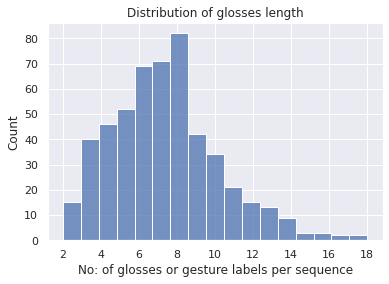} }
    \subfigure[Test data]{\includegraphics[width=0.3\textwidth]{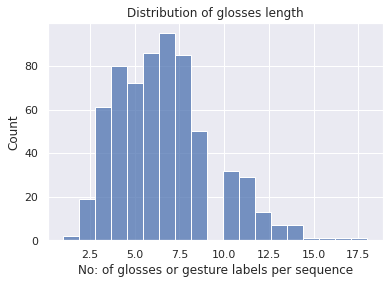}}
    \caption{Distribution of gloss lengths in train/validation/test sets of Phoenix2014T.  }
    \label{fig:gloss}
\end{figure}

The proffered ConSignformer approach was evaluated on the challenging German Sign Language datasets: RWTH-Phoenix-Weather-2014T (PHOENIX14T) \cite{camgoz2018neural} and RWTH-Phoenix-Weather-2014 (PHOENIX14) \cite{KOLLER2015108}. PHOENIX14T, an extension of the PHOENIX14 corpus, incorporating German translations for the video content, has a slightly reduced vocabulary than PHOENIX14 owing to enhancements in the normalization methods.  Consequently, while performance on PHOENIX14 and PHOENIX14T will exhibit similarities, they may not be directly comparable. The distribution of the glosses in train/val/test sets of Phoenix14T, based on the gloss lengths are represented in Figure \ref{fig:gloss}. The maximum sequence length is 28 in the training set and 18 in the validation set whereas it is 17 in the test set. In sequence-to-sequence tasks, the length of the sequence is a critical factor in decoding the sequence. 

\subsection{Implementation Details}
\begin{table}[htb]
\centering
\caption{\textbf{Configurations}}\label{tab:config}
\begin{adjustbox}{max width=0.46\textwidth}
\begin{tabular}{@{}p{40mm}@{} ll@{}p{30mm}@{}}
\toprule
\textbf{Hyper-parameter} & \textbf{Value}                                                 \\ \midrule
Number of encoders                 &    3 \\
Number of heads                      &  8 \\
Dropout                             & 0.1 \\
Learning rate                 & 0.001     \\
Number of epochs &              40        \\
Optimizer      & Adam \\
Batch size                   & 8 \\
Weight decay                  & 0.001 \\
Scheduler                     & cosineannealing \\
Expansion factor & 4 \\
Kernel size &  31 \\
 \midrule                                    
\end{tabular}
\end{adjustbox}
\end{table}

We commence with the unsupervised pretraining of Conformer on the pose dataset. This entails employing 3 encoder layers, each with a dimensionality of 512 and 8 attention heads. This pretraining is conducted as single-GPU training on the A100. Subsequently, we utilize the pretrained weights of S3D, available in PyTorch, and the pretrained Conformer weights for task-adaptive pretraining of the RGB and Heatmap Encoders. These pretrained weights are then employed in the final joint training of the ensemble, in conjunction with APNs, all under CTC supervision. The hyperparameter settings for the final training are detailed in Table \ref{tab:config}. This training is conducted on a single node, using multiple GPUs with 8 A100 GPU cards.

Glosses were decoded using a greedy search \cite{graves2012sequence} during both training and validation. However, during inference, the beam search decoding was employed, with beam sizes ranging from 0 to 10  \cite{boulanger2013audio}. To account for the length of the output, we applied a length penalty based on \cite{wu2016google}, where $\alpha$ ranged from 0 to 2. The optimal combination of $\alpha$ and beam size was determined through model evaluation on the development set and subsequently validated on the test set. The entire implementation is in PyTorch framework \cite{paszke2017automatic}.  Tensorflow \cite{abadi2016tensorflow} implementation was employed for CTC beam search decoding.

\subsection{Evaluation Metrics}
Word Error Rate (WER) is the metric that is used to evaluate the automatic sign recognition systems \cite{koller2015continuous}. It quantifies the difference between the recognized gloss sequence and the reference or ground truth sequence. WER is computed as follows:
\begin{equation}
    WER\ =\ \frac{\#substitutions\ \textbf{+}\ \#deletions\ \textbf{+}\ \#insertions}{\#words\ in\ hypothesis}.
\end{equation} 
The resulting WER value is usually expressed as a percentage, with lower values indicating better accuracy. 

\section{Experimental Results}

\begin{table*}[]
\centering
\caption{Performance results. (a) and (b) are the results of RGB/Heatmap encoders, trained separately. (c)-(g) shows the results of each network in the ensemble. The performance of the proposed network is given in (h). }
\label{results}
\begin{tabular}{@{}ll|ll|ll@{}}
\toprule
\multicolumn{2}{l}{} &
  \multicolumn{2}{c}{\textbf{Phoenix14T}} &
  \multicolumn{2}{c}{\textbf{Phoenix14}} \\ \midrule
\textbf{Sl. no:} &
  \textbf{Model} &
  \textbf{Dev WER} &
  \textbf{Test WER} &
  \textbf{Dev WER} &
  \textbf{Test WER} \\ \midrule
(a) &
  \begin{tabular}[c]{@{}l@{}}RGB Encoder\\ (Pretrained)\end{tabular} &
 20.8  & 21.62
 & 21.88  &  22.5
   \\ \midrule
(b) &
  \begin{tabular}[c]{@{}l@{}}Heatmap Encoder\\ (Pretrained)\end{tabular} &
 27.94  & 27.89
   &29.47
   &28.46
   \\ \midrule
\multicolumn{6}{l}{\textbf{Proposed Model Components}} \\ \midrule
(c)  & RGB Encoder                      & 18.52  & 19.60  & 19.24  & 19.54 \\ \midrule
(d)  & Heatmap Encoder                  &  18.92 &  19.90 & 19.33  & 19.65 \\ \midrule
(e)  & Fusion Network                   & 18.23  &  18.90 & 18.83  & 19.02 \\ \midrule
(f)  & RGB-APN                          & 20.1  &  21.51 & 20.99  & 20.50 \\ \midrule
(g)  & Heatmap-APN                      & 20.1 &   20.9 & 23.00  &22.72  \\ \midrule
(h)  & \textbf{ConSignformer (c+d+e)}  & 17.9  & 18.55  &  18.59 & 18.59 \\ \bottomrule
\end{tabular}
\vspace{0.5cm}

\end{table*}

The recognition results obtained by our models on the two benchmark datasets are presented in Table \ref{results}. Initially, we utilized the Conformer pretrained with Regressional Feature Extraction for task-adaptive CTC training of RGB and Heatmap encoders. The corresponding error rates are as illustrated in rows (a) and (b) respectively. It is noteworthy that the network utilizing raw videos performs significantly better compared to the one using heatmap inputs. Subsequent entries in the table demonstrate end-to-end training results. The performance disparities between RGB and heatmap inputs are evident from these values, although both exhibit improvement in the combined setup. The Fusion network, which processes fused data (RGB + heatmap), outperforms both individual inputs. The ensemble setup of the three networks, ConSignformer, displays superior recognition capabilities compared to the individual ones. The auxiliary pyramids network also demonstrates reasonably good results. Notably, the proposed model yields the best results in all the conducted experiments.
\begin{figure}[h]
    \centering
    \includegraphics[scale = 0.55]{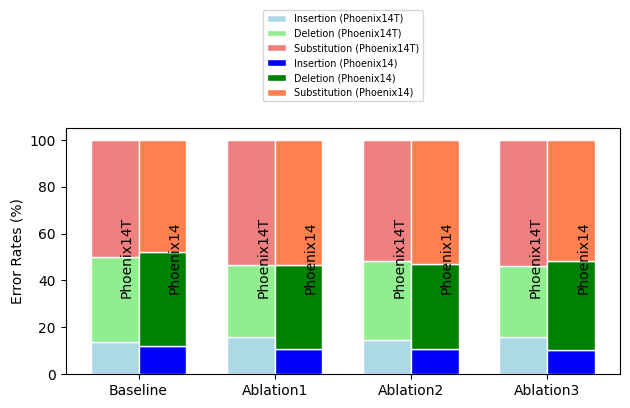}
    \caption{Variation in insertion, deletion, substitution error rates on Phoenix14 and Phoenix14T.}
    \label{ins}
\end{figure}

\subsection{Ablation Study}
We have conducted a comprehensive study by systematically ablating each of the major components of ConSignformer, including unsupervised pretraining, CMRA, and APNs, as denoted in Table \ref{ablation_models}. We initiated the ablation process by omitting all three components, forming the baseline model, which consists solely of the ensemble without the pretrained Conformer, CMRA, or APNs. Subsequently, we introduced the pretrained Conformer in one trial (Ablation 1), followed by the addition of CMRA in the next trial (Ablation 2), and in the final experiment (Ablation 3), we incorporated APNs. The Word Error Rates (WERs) for each of these trials are presented in Table \ref{ablation}. It is evident from the results that each module significantly contributes to the performance of ConSignformer.

Since WER is a combination of insertion, substitution, and deletion errors, it is worthwhile to conduct an in-depth study of the variation in these error values with the inclusion of each component. Figure \ref{ins} provides a visual representation of this study. Across models and datasets, insertion errors remain comparatively low and well-controlled. Substitution errors increase by 7.34\% and 8.34\%, while at the same time, the deletion rate is reduced by 17.16\% and 6.18\% from the baseline to the final model on Phoenix14T and Phoenix14, respectively. These results imply the proposed models' capability for more accurate and reliable sign recognition.

\begin{table*}
\centering
\caption{Ablation models }
\label{ablation_models}
\begin{tabular}{lccc}
\toprule
\textbf{Configuration} & \textbf{Pretrained Conformer} & \textbf{Cross-Modal Attention} & \textbf{Pyramids} \\
\midrule
Baseline &  & & \\
Ablation 1 &  $\checkmark$ & & \\
Ablation 2 & $\checkmark$ & $\checkmark$&  \\
Ablation 3 & $\checkmark$ & $\checkmark$ & $\checkmark$\\
\bottomrule
\end{tabular}
\vspace{0.5cm}
\caption{Ablation study results }
\label{ablation}
\begin{tabular}{lcc|cc}
\toprule
\multicolumn{3}{c}{\textbf{Phoenix14T}} & \multicolumn{2}{c}{\textbf{Phoenix14}} \\
\cmidrule(lr){1-3} \cmidrule(lr){4-5}
\textbf{Ablation} & \textbf{Dev WER} & \textbf{Test WER} & \textbf{Dev WER} & \textbf{Test WER} \\
\midrule
Baseline & 19.10 & 19.25 & 18.90 & 19.30 \\
Ablation 1 & 18.50  & 18.90 & 18.78 & 19.10 \\
Ablation 2 & 18.45 & 18.69 & 18.88 & 18.90\\
Ablation 3 &17.90  & 18.55 & 18.59  &18.59 \\
\bottomrule
\end{tabular}
\vspace{0.5cm}

\end{table*}
\begin{table}[]
\caption{Comparison with recent works on CSLR. The previous best results on the test sets are highlighted in yellow.}
\begin{adjustbox}{max width=0.48\textwidth}
\centering
\begin{tabular}{@{}llll|ll@{}}
\toprule
\multicolumn{1}{c}{} & \multicolumn{2}{c}{\textbf{Phoenix14T}} && \multicolumn{2}{c}{\textbf{Phoenix14}} \\ 
\cmidrule{2-3} \cmidrule{5-6}
\textbf{Model} & \textbf{Dev} & \textbf{Test} && \textbf{Dev} & \textbf{Test} \\
\midrule
SubUNets \cite{cihan2017subunets} &-  & - &&  40.80 & 40.70  \\
\midrule
LS-HAN \cite{huang2018video} &-  & - && - & -  \\
\midrule
IAN \cite{pu2019iterative} & - &-  &&  37.10 & 36.70  \\
\midrule
ReSign\cite{koller2017re} &-  & - &&  27.10 & 26.80  \\
\midrule
\begin{tabular}[c]{@{}l@{}}CNN-LSTM-HMMs \\ (Multi-Stream) \cite{koller2019weakly}\end{tabular} &22.10 & 24.10&& 26.00 & 26.00 \\
\midrule
SFL \cite{niu2020stochastic} &  25.10&26.10  && 24.90 & 25.30  \\
\midrule
DNF (RGB)\cite{cui2019deep}& - &-  && 23.80  & 24.40  \\
\midrule
FCN \cite{cheng2020fully}& 23.30 &25.10  && 23.70 & 23.90  \\
\midrule
DNF (RGB+Flow) \cite{cui2019deep}& - & - &&  23.10 & 22.90 \\
\midrule
Joint-SLRT \cite{camgoz2020sign}&24.60  & 24.50 &&  -&-  \\
\midrule
VAC \cite{min2021visual}&-  & - &&  21.20 & 22.30  \\
\midrule
LCSA \cite{zuo2022local}& - &-  &&  21.40 & 21.90  \\
\midrule
CMA \cite{pu2020boosting}& - &-  &&  21.30 & 21.90  \\
\midrule
SignBT \cite{zhou2021improving}& 22.70 &23.90  && - &-  \\
\midrule
GRU-RST \cite{aloysius2021incorporating}& 23.00 & 23.50 &&-  & - \\
\midrule
MMTLB \cite{chen2022simple}& 21.90 &22.50  && - &  -\\
\midrule
SMKD \cite{hao2021self}&  20.80 & 22.40  &&  20.80 & 21.00  \\
\midrule
STMC-R (RGB+Pose) \cite{zhou2021spatial}&  19.60 & 21.00  &&  21.10 & 20.70  \\
\midrule
C2SLR (RGB+Pose) \cite{zuo2022c2slr}&  20.20 & 20.40  &&  20.50 & 20.40  \\
\midrule
TwoStream-SLR \cite{chen2022two}&  17.70 & \hl{19.30}  &&  18.40 & \hl{18.80}  \\
\midrule
\textbf{\begin{tabular}[c]{@{}l@{}}ConSignformer \\ (Proposed Model)\end{tabular}} & 17.90 & \textbf{18.55} && 18.59 & \textbf{18.59} \\
\bottomrule
\end{tabular}
\end{adjustbox}
\vspace{0.5cm}

\label{sota}
\end{table}

\subsection{Comparison with State-of-the-art Architectures on CSLR}
In the context of CSLR, we conducted a comparison between our ConSignformer and state-of-the-art architectures on the Phoenix-2014 and Phoenix-2014T datasets, as detailed in Table \ref{sota}. Our recognition network has achieved a new state-of-the-art performance on both datasets, surpassing the previous best method by 1.12\% on Phoenix-2014 and by 3.88\% on Phoenix-2014T. We firmly believe that this outstanding performance can be attributed primarily to the superior context learning capabilities of the Conformer model.

\section{Discussion}

Our proposed ConSignformer surpasses the performance of the best-known Transformer, LSTM-based model, or a similarly sized convolutional model on the test sets of both Phoenix14 and Phoenix14T. To the best of our knowledge, this is the first work to adapt the Conformer architecture for a computer vision-based task. Our results clearly demonstrate the effectiveness of combining Transformer and convolution in a single neural network, as in the Conformer, particularly in the context of sign recognition. The utilization of swish activations contributed to faster convergence in the Conformer-based model.

When examining pretraining methods for Transformers and their variants, Masked Region Reconstruction and Contrastive Loss are widely adopted unsupervised tasks within the realm of computer vision. In our case, we introduce a novel approach by using the corrupted body keypoints vector. The Conformer is trained in an unsupervised manner to predict the clean vector by minimizing the mean squared error. Since it involves predicting a value, it can be considered as a regression task. However, since our objective is to obtain a clean feature vector, we refer to it as 'Regressional Feature Extraction.' The ablation study clearly demonstrates the advantages of this pretraining method. It is important to note that our pretraining is conducted using keypoints data from ISL (Indian Sign Language) videos, while our fine-tuning is performed on a German dataset. This highlights the versatility of our pretrained Conformer model, which can be effectively employed for various sign recognition tasks, regardless of the nature of the sign language being used.

Another factor that has contributed to the performance of ConSignformer is CMRA. While several related studies have explored various forms of cross-modal attention, our work is the first to adapt the relative self-attention mechanism of the Conformer model into a cross-modal type. This modification has proven to be highly effective, both in the ensemble setup and in the final proposed model.

Each individual network in our proposed system consists of a CNN and a Conformer, which creates a deep and complex model that is susceptible to issues like vanishing gradients during training. To mitigate such challenges when training deep networks, the addition of auxiliary tasks is effective for the primary tasks. Prior state-of-the-art work \cite{chen2022two} in CSLR has employed pyramid networks as feature extractors for the auxiliary task of CTC. We have adopted a similar approach for our auxiliary task, introducing a novel pyramid network that leverages attention weights to learn the optimal combination of features. This method has proven to be more beneficial than simply combining all features.

The ablation study reveals the impact of each component of ConSignformer. Unlike other recent works, a detailed analysis of insertion, substitution, and deletion rates has been lacking in the literature. Our study demonstrates a substantial reduction in the deletion rate across various model configurations and datasets, indicating improved performance. This reduction in deletion errors suggests that the models are becoming more accurate in recognizing and retaining signed words. Furthermore, the fact that insertion errors remain consistently low and controlled is a positive indicator. It implies that the models do not overcompensate for the reduction in deletion errors by introducing excessive insertion errors, which could lead to inaccurate transcriptions. Although substitution errors increased slightly, the overall effect on WER is limited. These positive changes are consistent across both datasets, demonstrating the models' generalizability in CSLR tasks.

\textbf{Limitations:} Despite achieving state-of-the-art results with the adapted Conformer model, its practical applicability in real-world scenarios is hindered by the associated resource costs. Training in a multi-GPU setup can be prohibitively expensive, and once trained, the model's inference time may impose limitations on certain devices or specific use cases. Therefore, designing a more efficient architecture is crucial for creating a production-ready system that is both lightweight and capable of fast decoding. This is in line with the emerging research on Accessible and Inclusive Artificial Intelligence (AIAI) \cite{kinnula2021researchers} where inclusivity and efficiency are key concerns. Our future work will focus on this direction — designing a production-ready model that is less resource-intensive while maintaining efficiency, thus contributing to the broader goals of AIAI and ensuring the accessibility and usability of sign language recognition technology.

\section{Conclusion}
In this study, we introduced ConSignformer, a model composed of ensembles of S3D and Conformer, tailored for sign recognition. The Conformer is pretrained using a novel unsupervised method known as Regressional Feature Extraction. Within the ensemble setup, the Conformer incorporates Cross-Modal Relative Attention, designed specifically for this purpose. The training of ConSignformer is enhanced by Attentional Pyramid Networks, serving as auxiliary tasks. This model achieves state-of-the-art performance on the benchmark German datasets, Phoenix14 and Phoenix14T. However, the model's complexity poses challenges for practical real-world applications. Therefore, our future work will focus on designing an efficient, production-ready model.

 \bibliographystyle{IEEEtran}
\bibliography{refs}

\mycomment{
\begin{figure*}[!t]
\centering
\subfloat[]{\includegraphics[width=2.5in]{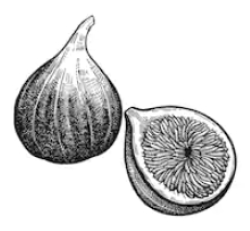}%
\label{fig_first_case}}
\hfil
\subfloat[]{\includegraphics[width=2.5in]{fig1}%
\label{fig_second_case}}
\caption{Dae. Ad quatur autat ut porepel itemoles dolor autem fuga. Bus quia con nessunti as remo di quatus non perum que nimus. (a) Case I. (b) Case II.}
\label{fig_sim}
\end{figure*}
}
\section*{Acknowledgments}
This project draws inspiration and guidance from Amma, the Chancellor of Amrita University. We wish to extend our heartfelt appreciation to the Ministry of Electronics and Information Technology (Meity), Government of India, for their generous financial support towards this research. 

\end{document}